\documentclass[letterpaper, 10 pt, conference]{ieeeconf}  

\IEEEoverridecommandlockouts                              

\usepackage[english]{babel}
\usepackage{subfigure}
\usepackage{bm}
\usepackage{subfig}
\usepackage{amsmath}
\usepackage{tensor}
\usepackage{amsfonts}
\usepackage{algorithm}
\usepackage{algorithmic}
\usepackage{url}
\usepackage{amssymb}
\usepackage{graphicx}
\usepackage{amsmath}
\usepackage{breqn}
\usepackage{siunitx}
\usepackage{xcolor}
\usepackage{afterpage}
\usepackage{soul}
\usepackage{tabularx}
\usepackage{subfigure}
\usepackage{multirow}
\usepackage{here} 
\usepackage{relsize}
\newcolumntype{x}{>{\hsize=.5\hsize}X}

\title{\LARGE \bf Aggressive Visual Perching  with Quadrotors on Inclined Surfaces}

\author{Jeffrey Mao$^{1}$, Guanrui Li$^{1}$, Stephen Nogar$^{2}$,  Christopher Kroninger$^{2}$, and Giuseppe Loianno$^{1}$ 
\thanks{$^{1}$The authors are with the New York University, Tandon School of Engineering, Brooklyn, NY 11201, USA
        {\tt\footnotesize email: \{jm7752, lguanrui, loiannog\}@nyu.edu}.}%
\thanks{$^{2}$The authors are with the U.S. Army Research Laboratory, 2800 Powder Mill Road, Adelphi, MD 20783, USA.
            {\tt\footnotesize email: \{stephen.m.nogar, christopher.m.kroninger\}.civ@mail.mil.}}
\thanks{This work was supported by the ARL grant DCIST CRA W911NF-17-2-0181 and the young researchers program "Rita Levi di Montalcini" 2017 grant PGR17W9W4N.}
}

\begin{document}

\maketitle
\thispagestyle{empty}
\pagestyle{empty}

\begin{abstract}
Autonomous Micro Aerial Vehicles (MAVs) have the potential to be employed for surveillance and monitoring tasks. By perching and staring on one or multiple locations aerial robots can save energy while concurrently increasing their overall mission time without actively flying. In this paper, we address the estimation, planning, and control problems for autonomous perching on inclined surfaces with small quadrotors using visual and inertial sensing. We focus on planning and executing dynamically feasible trajectories to navigate and perch to a desired target location with on board sensing and computation. Our planner also supports certain classes of nonlinear global constraints by leveraging an efficient algorithm that we have mathematically verified. The on board cameras and IMU are concurrently used for state estimation and to infer the relative robot/target localization.  The proposed solution runs in real-time on board a limited computational unit. Experimental results validate the proposed approach by tackling  aggressive perching maneuvers with flight envelopes that include large excursions from the hover position on inclined surfaces up to 90$^\circ$, angular rates up to 600~deg/s, and accelerations up to 10~m/s$^2$.



\end{abstract}

\section{Introduction}

Micro Aerial Vehicles (MAVs) have great speed and maneuverability however they tend to have very low flight time. 
Current solutions limit battery life to around 10-20 minutes. Fortunately for many missions such as environmental monitoring, it is unnecessary to remain in hover for the whole mission duration. In general, by perching and staring on one or multiple location, a MAV can greatly extend its mission time saving power without the need to frequently replace batteries. This motivates the need of autonomous perching solutions to conserve energy and extend the mission time.

\begin{figure}
    \centering
    \includegraphics[width=1\columnwidth]{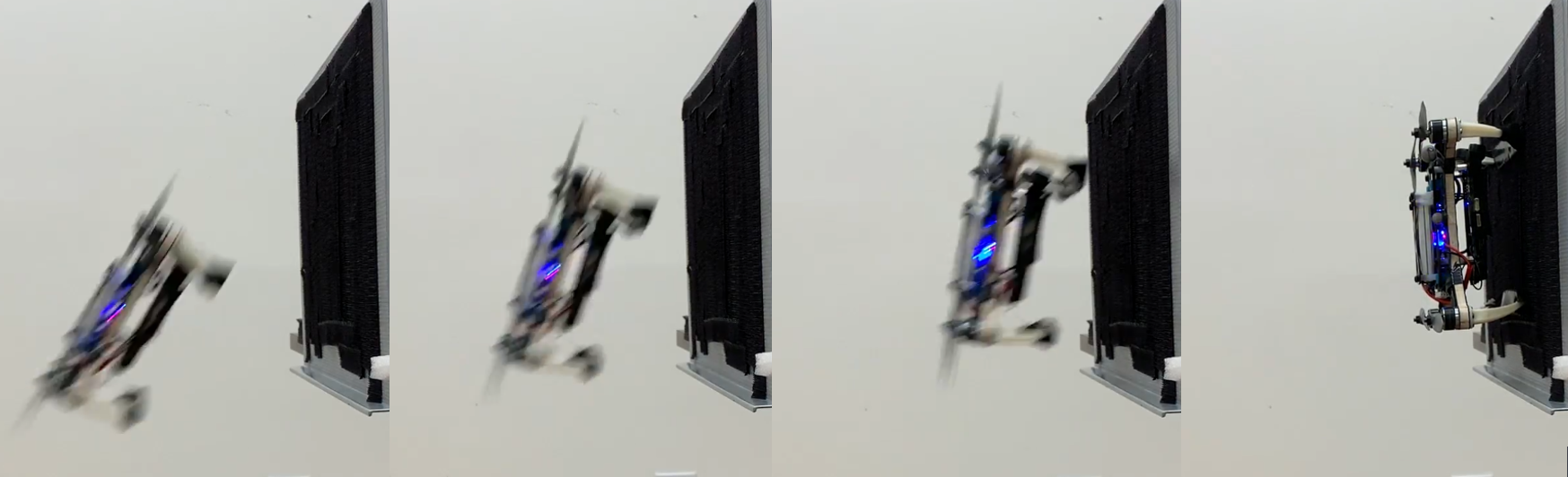}
    \caption{Aggressive visual perching sequence maneuver for a 90$^\circ$ inclined surface.}
    \label{fig:90_perchingimage}
    \vspace{-5mm}
\end{figure}

In this paper, we tackle the autonomous perching problem on inclined surfaces with quadrotors solely using on board cameras and Inertial Measurement Unit (IMU) as shown in Fig.~\ref{fig:90_perchingimage}. Inclined flat surfaces like walls and rooftops are plentiful especially in urban environments and by focusing on this avenue, we can aim to greatly reduce the energy consumption of multiple types of mission. The proposed autonomous perching problem is challenging for several reasons. The maneuver, to intercept the target, requires large excursions from the hover position. In addition, the vehicle must generate and execute dynamically feasible trajectories respecting the actuator and sensor constraints despite the presence of nonholonomic and underactuation constraints. Finally, in our case, the maneuver has to be accomplished relying exclusively on on board minimalist sensor data (camera and IMU) and limited computational unit. The ability to execute these challenging maneuvers can be leveraged as well in several other scenarios including reaction to sudden changes in the operational conditions for obstacle avoidance or navigation in constrained environments.

This paper presents multiple contributions. First, we show how to generate and execute dynamically and physically feasible trajectories for perching on inclined surfaces. Our planning solution efficiently supports certain classes of nonlinear constraints such as maximum thrust limit through the use of an efficient bound checking algorithm that we have mathematically verified. The planner versatility to incorporate a diverse set of constraints makes it potentially able to support different perching or adhesion mechanisms. Second, our approach relies solely on on board sensing and computation for navigation and to infer the relative robot/target configuration. Finally, this is the first time that a fully autonomous quadrotor system can perch on any flat inclined surface with minimum mechanical modifications. Other works either assume the availability of a motion capture system~\cite{Thomas_perch_inclined} or require the landing point to be in the field of view at the starting location based on visual servoing approaches~\cite{Thomas_visual_servoing,Zhang_optimal_traj}.


\begin{figure*}[!t]
    \centering
    \includegraphics[width=\textwidth]{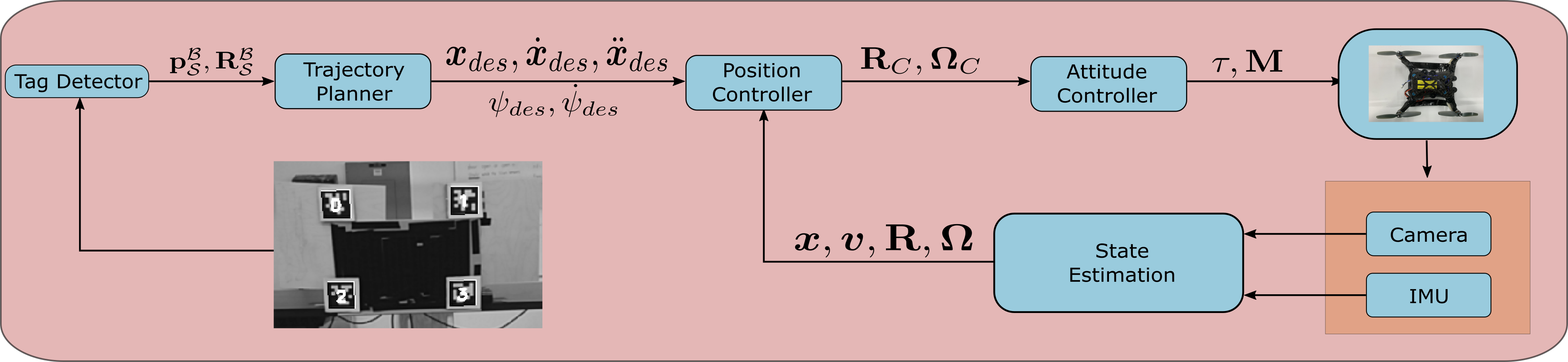}
    \caption{System architecture for the perching task}
    \label{fig:blockdiagram}
    \vspace{-10pt}
\end{figure*}

\section{Related Works}~\label{sec:related_works}
Prior works on perching with quadrotors on inclined surfaces focus on solving the trajectory generation and/or control problems~\cite{Mellinger2012, Thomas_perch_inclined} relying exclusively on motion capture systems. These solutions do not address the design challenges and requirements when deploying robots equipped with embedded on board sensors and computationally limited units. Furthermore, the approach presented in~\cite{Mellinger2012} relies on composition of multiple control modes with linearized controllers without guaranteeing feasibility of the maneuver. Other works focus on the perching mechanism design. The approach proposed in~\cite{Chi2014} \cite{AGRASP} use claws which limits perching to cylindrical objects of the appropriate width to fit the claw's grip. Other solutions rely on dry adhesives~\cite{Kalantari2015,Hawkes2013,Daler2013}, suction gripper solutions~\cite{Tsukagoshi2015,Kessens2016} mechanisms, or active perching mechanisms solutions~\cite{Tsukagoshi2015}. However, the use of active perching mechanism solutions further increases the vehicle's payload and energy requirements, while concurrently decreasing the overall flight time. Finally, in~\cite{AGRASP} a bio-inspired trajectory planning approach is presented. However, the work relies on an actuated gripper for specific cylindrical objects and relies on a heavy 3D camera to localize perching targets. Finally, visual servoing approaches~\cite{Thomas_visual_servoing,Zhang_optimal_traj} have  shown autonomous perching results without the use of motions capture but are highly dependent on objects' shapes and require the object to initially be in the field of view.

Other works employ fixed wing solutions and focus on perching mechanism designs~\cite{Moore_2014,Lussier2011}. Fixed wing solutions have lower maneuverability with respect to quadrotor solutions. Moreover, a quadrotor can hover in place and navigate in confined environments with both slow and fast agile movements. The flight time is more restricted with respect to fixed wing solutions further motivating the usefulness to provide autonomous perching solutions for quadrotors.

Compared to the aforementioned solutions, we concurrently guarantee dynamical and physical feasibility of the planned trajectories. Furthermore, our approach relies exclusively on on board sensing and estimation with a small computational unit. In the presented case, we focus on guaranteeing on board autonomous perching with dynamically and physically feasible trajectories. We leverage the differential flatness property and develop an efficient planning algorithm to generate trajectories for a quadrotor without the limitation to first order systems~\cite{Zhang2013} or composition of multiple control modes relying on linearized controllers~\cite{Mellinger2012}.  The proposed approach is agnostic with respect to the type of active/passive 
perching mechanism, due to the ability to support a diverse set of constraints which could conform to different attachment/adhesion mechanisms.

\section{System Overview}~\label{sec:overview}
The proposed system architecture is shown in Fig.~\ref{fig:blockdiagram}
is a quadrotor running with a Qualcomm\textsuperscript{\textregistered} Snapdragon\textsuperscript{TM} board and 4 brushless motors. The Qualcomm\textsuperscript{\textregistered} Snapdragon\textsuperscript{TM} has a Qualcomm\textsuperscript{\textregistered} Hexagon\textsuperscript{TM} DSP, Wi-Fi, Bluetooth, GPS, one core processor along with a downward and  front-facing camera with a $160^\circ$ field of view along with an IMU. For perching, we employ VELCRO\textsuperscript{\textregistered} material mounted in the ventral part of the vehicle.

The software framework has been developed in ROS\footnote{\url{www.ros.org}} on a Linux kernel and includes a state estimation algorithm running at 500 Hz composed of a Unscented Kalman Filter (UKF) and Visual Inertial Odometry (VIO)~\cite{loianno_est} which processes images at 30 Hz. The downward facing camera is solely devoted to state estimation of the quadrotor and the front-facing camera is used to detect the perch target's position and orientation. Furthermore, the system runs on board a position and attitude controllers plus a trajectory planner to generate and execute planned path.


\section{Approach}~\label{sec:approach}
\begin{figure}[t]
    \centering
    \includegraphics[width=\linewidth]{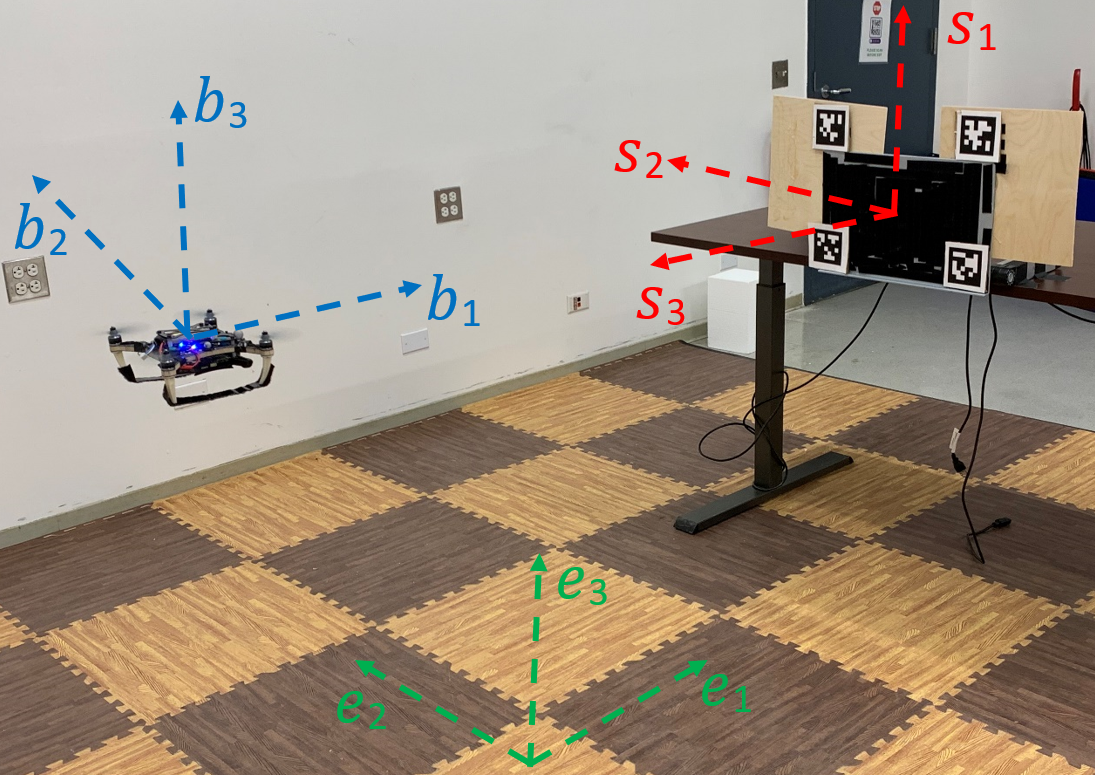}
    \caption{Setup overview and frame convention definitions.}
    \label{fig:Perching Setup}
        \vspace{-10pt}
\end{figure}

Let the inertial frame $\mathcal{I}$ be represented by the following three axes $\begin{bmatrix} \mathbf{e}_1 & \mathbf{e}_2 & \mathbf{e}_3 \end{bmatrix}$. The quadrotor body frame $\mathcal{B}$ is represented by $\begin{bmatrix} \mathbf{b}_1 & \mathbf{b}_2 & \mathbf{b}_3 \end{bmatrix}$. This frame origin is located at the center of mass of the vehicle. Consequently, we denote the relative position of the quadrotor with respect to the $\mathcal{I}$ frame as $\bm{x} = \begin{bmatrix} x & y & z \end{bmatrix}^{\top}$ and the relative orientation as $\mathbf{R}= \begin{bmatrix} \mathbf{b}_1 & \mathbf{b}_2 & \mathbf{b}_3 \end{bmatrix} \in SO(3)$.  The perching target frame is denoted with $\mathcal{S}$ and is represented by the axes $\begin{bmatrix} \mathbf{s}_1 & \mathbf{s}_2 & \mathbf{s}_3 \end{bmatrix}$. The relevant frames and configuration settings are depicted in Fig.~\ref{fig:Perching Setup}. The perching problem from time $t=t_0$ to $t=t_f$  requires the vehicles to navigate by planning and executing a feasible trajectory  (i.e., generating a sequence of $\mathbf{R}\left(t\right)\in SO(3)$ and $\bm{x}\left(t\right)\in {\mathbb{R}} ^3$) such that $\mathcal{B}\equiv\mathcal{S}$ at $t=t_f$. The problem is decomposed in several steps. First, the vehicle visually locates the target and estimates the relative configurations from the $\mathcal{B}$ to $\mathcal{S}$ frames (i.e, relative position $\mathbf{p}_{\mathcal{S}}^{\mathcal{B}}\in\mathbb{R}^3$ and orientation $\mathbf{R}_{\mathcal{S}}^{\mathcal{B}}\in SO(3)$). Second, the relative configuration information is incorporated at the planning and control levels to generate and execute trajectories that are dynamically and physically feasible.

In Section~\ref{sec:modelcontrol}, we briefly review the vehicle's system dynamics and control. Section~\ref{sec:estimation} describes the proposed perception pipeline including autonomous navigation and target localization. Section~\ref{sec:planning} details our trajectory planning approach and how we leverage in this context the system dynamics and the relative configuration constraint to guarantee the correct plan and execution.

\subsection{Modeling and Control}\label{sec:modelcontrol}
The system dynamic model in the inertial frame $\mathcal{I}$ is 
\begin{equation}\label{eqn:model}
\begin{split}
&\dot{\bm{x}} = \bm{v}, \dot{\bm{v}} = \boldsymbol{a}, m\boldsymbol{a} = \mathbf{R}\tau \mathbf{e}_3 - mg\mathbf{e}_3, 
\\
&\Dot{\mathbf{R}} = \mathbf{R}\hat{\bm{\Omega}}, \mathbf{J}\Dot{\bm{\Omega}} + \bm{\Omega}\times \mathbf{J}\bm{\Omega} = \mathbf{M},
\end{split}
\end{equation}
where $\bm{x}, \bm{v}, \boldsymbol{a} \in {\mathbb{R}} ^3$ are the position, velocity, acceleration of the quadrotor's center of mass in Cartesian coordinates with respect to the inertial frame $\mathcal{I}$, $\mathbf{R}$ represents the orientation of the quadrotor with respect to $\mathcal{I}$.  $\mathbf{\Omega} \in {\mathbb{R}} ^3$ is the angular velocity of the quadrotor with respect to $\mathcal{B}$, $m\in\mathbb{R}$ denotes the mass of the quadrotor, $\mathbf{J} \in {\mathbb{R}}^{3 \times 3}$ represents its inertial matrix with respect to $\mathcal{B}$, $g = 9.81 m/s^2$ is the standard gravitational acceleration, $\mathbf{M}\in\mathbb{R}^3$ is the total moment with respect to $\mathcal{B}$, $\tau\in {\mathbb{R}} $ represents the total thrust to the quadrotor, and the $\hat{\cdot}$ represents the mapping such that $\hat{\mathbf{a}}\mathbf{b} = \mathbf{a} \times \mathbf{b}, \forall \mathbf{a},\mathbf{b} \in \mathbb{R}^{3}$.

To achieve aggressive maneuvers, we apply a nonlinear geometric controller that was leveraged in our previous work~\cite{LoiannoRAL2017} to achieve agile flight in indoor environments. First, $\mathbf{k}_R, \mathbf{k}_{\Omega}, \mathbf{k}_x, \mathbf{k}_{v}\in\mathbb{R}^{3\times3}$ are positive definite diagonal matrix representing the feedback gains for the errors in orientation, angular velocity, position and velocity respectively. Based on those feedbacks, the control inputs are thrust $\tau$ and moment $\mathbf{M}$ selected as

\begin{equation}\label{eqn:control}
\begin{split}
\tau &= \left(-\mathbf{k}_x \mathbf{e}_x - \mathbf{k}_{v} \mathbf{e}_{v} + mg\mathbf{e}_3 + m \ddot{\bm{x}}\right) \cdot \mathbf{R}\mathbf{e}_3=\mathbf{f}\cdot \mathbf{R}\mathbf{e}_3, \\
\mathbf{M} &= -\mathbf{k}_R \mathbf{e}_R - \mathbf{k}_{\Omega} \mathbf{e}_{\Omega} + \bm{\Omega} \times \mathbf{J}\bm{\Omega}  \\ &\hspace{50pt}-\mathbf{J}\left(\hat{\bm{\Omega}}\mathbf{R}^{\top}\mathbf{R}_C\bm{\Omega}_C - \mathbf{R}^{\top}\mathbf{R}_C\dot{\bm{\Omega}}_C\right).
\end{split}
\end{equation}
$\mathbf{e}_R, \mathbf{e}_{\Omega}, \mathbf{e}_x, \mathbf{e}_v\in\mathbb{R}^3$ are the orientation, angular velocity, position and velocity errors this is detailed in works ~\cite{McLamrock, Loianno_IJRR2018}, and the $*_C$ are the values obtained from the differentially flat outputs.  $*_{des}$ represents the differential flat outputs of the quadrotor system computed using the planning algorithm. 

\subsection{State Estimation and Perching Target Localization}\label{sec:estimation}

For quadrotor autonomous navigation, we leverage our previous work~\cite{loianno_est}, where we showed aggressive maneuvers combining visual and inertial data via VIO and UKF.

The perching localization method is unimportant for our system as long as it gives both position and orientation. For our experiments perching localization is achieved through the use of four Apriltags \cite{olson2011tags} placed on the four corners of the target. By placing these Apriltags on the corners, we can easily calculate the exact center of the landing pad by averaging these position while also leaving a large portion of VELCRO\textsuperscript{\textregistered} to land on. Our software perception pipeline polls images from the front camera and runs an Apriltag localization algorithm. The quadrotor polls the front-facing camera till it recognizes all 4 Apriltags Ids, and takes the first sample image as the target configuration location $\mathbf{p}_{\mathcal{S}}^{\mathcal{B}}$ and orientation $\mathbf{R}_{\mathcal{S}}^{\mathcal{B}}$. These target configurations are represented in the $\mathcal{I}$ frame. To reduce sensitivity to the noise from the detection process, we average the tags' position and quaternions to obtain the target center position and orientation.

\subsection{Planning for Aggressive Perching}\label{sec:planning}

After acquiring the target location and orientation, we plan a differentially smooth and dynamically feasible trajectory to satisfy the end goal of reaching the target and perching by exploiting the differential flatness property of the quadrotor. This allows us to shift the planning problem defined at the beginning of this Section to the flat space of the vehicle $\{\bm{x},\psi\}=\{x,y,z,\psi\}$, where $\psi $ is the yaw angle. We employ a set of polynomial splines $P_d$ to represent the quadrotor trajectory 
\begin{equation} \label{eqn:poly_spline}
\begin{split}
P_d(t) &= 
     \begin{cases}
       \text{$p_{1d}\left(t-t_0\right)$} &\quad\text{if $t 	\in \left[t_0,t_1\right]$}\\
        \text{$p_{2d}\left(t-t_{1}\right)$}&\quad\text{if $t 	\in \left[t_{2},t_{1}\right]$}\\
        \hspace{10pt}\vdots\\
        \text{$p_{fd}\left(t-t_{f-1}\right)$} &\quad\text{if $t 	\in \left[t_{f-1},t_{f}\right]$}\\
     \end{cases},\\
 p_{id}(t) &= \sum_{n=0}^{N} c_{nid}t^{n}~, i = 1, \cdots,f
     \end{split}
\end{equation}
where $f$ is the number of splines, $N$ is the polynomial order, and $d\in\{1,2,3, 4\}$ corresponds to the dimensions of the flat space of the quadrotor system composed by the flat outputs $\{x,y,z,\psi\}$, $p_{id}$ represents the $i^{th}$ polynomial making up the full trajectory of $P_d$, $c_{nid}\in\mathbb{R}$ is the $n^{th}$ coefficient of $p_{id}$. Polynomials are ideal because the function and its derivatives can be written as matrix multiplication. 

In the following, we formulate the trajectory planning problem as a Quadratic Programming (QP). 
The trajectory optimization must satisfy perching, actuators, and sensing constraints globally in a time efficient manner. 
We first declare a cost function as the squared norm of the $j^{th}$ order derivative summed in all dimensions. We formulate the case for one polynomial spline in the form
\begin{equation}\label{eqn:cost_integral_QP} \mathlarger{\min_{\mathbf{c}_{id}}} \hspace{10pt}\mathbf{c}_{id}^{\top} \left(
\int_{t_{i-1}}^{t_i} \frac{d^j \mathbf{t}_i}{dt^j} \frac{d^j \mathbf{t}_i}{dt^j}^{\top} \,dt \right)\mathbf{c}_{id}  = \mathlarger{\min_{\mathbf{c}_{id}}}~ \mathbf{c}_{id}^\top \mathbf{Q}_i \mathbf{c}_{id}
\end{equation}
where $\mathbf{t}_i= \left[1,(t-t_{i-1}),(t-t_{i-1})^2,\cdots(,t-t_{i-1})^N\right]^{\top}\in\mathbb{R}^{N}$ represents the time vector and and $\mathbf{c}_{id}\in\mathbb{R}^{N}$ is the vector that consists of all the coefficients of $p_{id}$. The term $\mathbf{Q}_i\in\mathbb{R}^{N\times N}$ is the cost matrix formed from the center integral in eq.~(\ref{eqn:cost_integral_QP}). It should be noted that $\mathbf{Q}_i$ is identical for all dimensions, $d$, so it has no subscript $d$. We can then convert eq.~(\ref{eqn:cost_integral_QP}) into one unified cost for dimension $d$ in the form $\mathbf{c}_d^T \mathbf{Q} \mathbf{c}_d$ by stacking the matrices diagonally Diag$(\mathbf{Q}_{1},\cdots, \mathbf{Q}_{i},\cdots, \mathbf{Q}_{f})=\mathbf{Q}$ and forming $\mathbf{c}_d\in \mathbb{R}^{Nf}$ as a vector of all the $\mathbf{c}_{id}$ stacked vertically. We can then formulate an equality constraint for endpoints of each spline in the form $\mathbf{b}_{id} = \mathbf{A}_{id} \mathbf{c}_{id}$. The term $\mathbf{b}_{id}$ is a vector consisting of all the user defined constraints for dimension $d$ that are imposed at time, $t_i$. $\mathbf{A}_{id}$ is defined by transposing $\mathbf{t}_i$ into a row vector then stacking $\mathbf{t}_i$ and its derivatives vertically as
\begin{equation}
\label{eqn:end_point_constr}
\begin{bmatrix}
p_{id}\left(t_i\right) \\
\Dot{p}_{id}\left(t_i\right) \\
\vdots \\

\end{bmatrix} = \begin{bmatrix}
 {\mathbf{t}_i\left(t_i\right)}&  {\Dot{\mathbf{t}}_i}\left(t_i\right) &  
 \hdots & 
\end{bmatrix}^{\top}*
\mathbf{c}_{id} .
\end{equation}
 The constraints on higher order terms like velocity and accelerations are optional to declare in eq.~(\ref{eqn:end_point_constr}). We can simply stack matrices in eq.~(\ref{eqn:end_point_constr}) diagonally,  Diag$(\mathbf{A}_{1d},\cdots, \mathbf{A}_{id},\cdots, \mathbf{A}_{fd})$, to create the combined constraint for all trajectories in a dimension. 
Additionally, a continuity constraint between the endpoints of all splines is enforced for all derivatives. The first order derivative case is  
\begin{equation}\label{eqn:continuity}
\resizebox{.9\hsize}{!}{$\begin{bmatrix}
0 \\
0 \\
\vdots \\
0\\
\end{bmatrix} = \begin{bmatrix}
 \Dot{ \mathbf{t}}_1(t_1) &  - \Dot{\mathbf{t}}_2(t_1) & 0 & ... \\
0 &   \Dot{ \mathbf{t}}_2(t_2) &  - \Dot{ \mathbf{t}}_3(t_2) & ...  \\
\vdots & \ddots & \ddots &  \vdots\\
0 & ... &   \Dot{ \mathbf{t}}_{f-1}(t_{f-1}) &  - \Dot{ \mathbf{t}}_f(t_{f-1}) 

\end{bmatrix}\begin{bmatrix}
\mathbf{c}_{1d} \\
\mathbf{c}_{2d} \\
\vdots \\
\mathbf{c}_{fd}\\
\end{bmatrix}.$}
\end{equation}
 Eq.~(\ref{eqn:continuity}) constraint is replicated to the fourth order and combined with eq.~(\ref{eqn:end_point_constr}) constraints by stacking the matrices vertically creating a unified equality constraint matrix $\mathbf{A}_d$.

We can add additional inequality constraints formatted the same as eq.~(\ref{eqn:end_point_constr}) $\mathbf{A}_{id}$ for each spline and combine them in the same way to get $\mathbf{G}_d$. The combined constraint and cost are solved as $4$ separate QP optimization in parallel for each dimension to speed up the overall solution time. For each QP dimension the optimization problem is detailed below
\begin{equation}
\begin{aligned}
\min_{\mathbf{c}_d} \quad & \mathbf{c}_d^T \mathbf{Q} \mathbf{c}_d\\
\textrm{s.t.} \quad & \mathbf{A}_d \mathbf{c}_d=\mathbf{b}_d\\
& \mathbf{y}_d \leq \mathbf{G_d} \mathbf{c}_d \leq \mathbf{z}_d    \\
\end{aligned}
\end{equation}
where the inequality constraint are between a lower bound $\mathbf{y}_d$  and upper bound $\mathbf{z}_d$ respectively. In the following section, we will discuss the perception, state and actuator constraints specific to executing perching tasks. 

\subsubsection{Perching Perception and Physical Constraints}
Inspired by~\cite{loianno_est}, we exploit the differential flatness property of our model to derive a relationship between the vehicle's center of mass acceleration, $\Ddot{\bm{x}}$ and rotation matrix $\mathbf{R}$. First we look at the dynamical model established in eq.~(\ref{eqn:model}). We can derive that the nominal thrust as
\begin{equation}\label{eqn:thrust}
 \tau  = m||\Ddot{\bm{x}} +g\mathbf{e}_3||. 
\end{equation}
Based on the nonholonomic  property of a quadrotor system, the generated force is along the $\mathbf{b}_3$ axis of the body frame of a quadrotor; therefore, the $\mathbf{b}_3$ should  satisfy
\begin{equation}\label{eqn:b3}
\mathbf{b}_3  = \frac{\Ddot{\bm{x}} +g\mathbf{e}_3}{||\Ddot{\bm{x}} +g\mathbf{e}_3|| }.
\end{equation}
Using eq.~(\ref{eqn:b3}), we can define $\mathbf{b}_3$ at the end of the perching at $t_f$ through an acceleration constraint as
\begin{equation} \label{eqn:acceleration constraint}
\Ddot{\bm{x}}(t_f) = \alpha\mathbf{s}_3 -g\mathbf{e}_3,
\end{equation}
where $\alpha=||\Ddot{\bm{x}}(t_f) +g\mathbf{e}_3||\in\mathbb{R}$ corresponds to the pre-defined thrust of the quadrotor selected by the user. The planned $\Ddot{\bm{x}}$ is then injected in the thrust in eq.~(\ref{eqn:control}). The $\mathbf{s}_3$ direction on the inclined surfaces is extracted from the last column of the matrix $\mathbf{R}_{\mathcal{S}}$. This orientation $\mathbf{R}_{\mathcal{S}}$ is obtained combining the outcome of the state estimation algorithm with the target detector, which also provides $\mathbf{p}_{\mathcal{S}}$. These quantities allow us to fully define the target configuration with respect to the inertial frame and provide the constraints required to accomplish the perching maneuver. 
To enforce this maneuver at control level, the rotation matrix, we impose that in eq.~(\ref{eqn:control}), $\mathbf{R}_C$ should be chosen according to
\begin{equation}
\mathbf{R}_C= \begin{bmatrix} 
\mathbf{b}_{1,C} & \mathbf{b}_{2,C} & \mathbf{b}_{3,C}
\end{bmatrix}
\end{equation}
\begin{equation}
\begin{split}
  \mathbf{b}_{1,C} =& \frac{\mathbf{b}_{2,des} \times \mathbf{b}_{3}}{\vert\vert{\mathbf{b}_{2,des} \times \mathbf{b}_{3}}\vert\vert},~~~
  \mathbf{b}_{2,C} = \mathbf{b}_{3} \times \mathbf{b}_{1},\\~\nonumber
    \mathbf{b}_{2,des} =& \begin{bmatrix} -\sin\psi_{des}, & \cos\psi_{des}, & 0 \end{bmatrix}^\top,~\nonumber
  \mathbf{b}_{3,C} = \frac{\mathbf{f}}{\vert\vert{\mathbf{f}}\vert\vert}.\nonumber
  \label{eq:R_des}
\end{split}
\end{equation}

The desired yaw angle $\psi_{des}$ can be chosen by the user. Generally, we select it such that $\mathbf{b}_{2,des}$ is parallel to $\mathbf{s}_2$, which can be know from $\mathbf{R}_{\mathcal{S}}$. The commanded angular rate is then
\begin{equation}
  \hat{\bm{\Omega}}_{C} = \mathbf{R}_{C}^\top\dot{\mathbf{R}}_{C}.
\end{equation}


Next, additional constraints have to be placed on the impact velocity to ensure that the quadrotor is neither moving too quickly nor too slowly so that it does not properly adheres to the perching mechanism. We take this aspect into account by imposing additional velocity constraints in the target proximity
\begin{equation}
v_{min} \leq \Dot{\bm{x}}(t_f) \cdot \mathbf{s}_3 \leq v_{max},
\label{eqn:Velocity constraint}
\end{equation}
where $v_{min}$ is the minimum impact velocity and $v_{max}$ is the maximum. 
Finally, the vehicle must complete most of its rotation before its impact with the surface rather than try to pivot at the target. This is to avoid the rim of the vehicle impacting the target surface. Should the vehicle begin rotating too late in its trajectory, the front end will collide with the landing pad before reaching the desired attitude. This aspect is expressed by enforcing an additional acceleration range by a given $q$ tolerance in proximity of the target.
In order to apply the inequality constraint to our optimization, we discretize the equation as
\begin{equation} \label{eqn:acceleration discrete ineq constraint}
\begin{split}
&\left(1-q\right)\left(\alpha\mathbf{s}_3 -g\mathbf{e}_3\right) \leq \Ddot{\bm{x}}(t) \leq \left(1+q\right)\left(\alpha\mathbf{s}_3 -g\mathbf{e}_3\right),
\\
&\forall  t \in{\{t_f-t_k+j*dt\}}  \hspace{10pt} j \in \mathbb{Z} \And 0 \leq  j < \frac{t_k}{dt},
\end{split}
\end{equation}
where $dt$ is the sampling time of our trajectory planner and $t_k$ is the time prior to the impact which is user defined. 

\subsubsection{Actuators Constraints}\label{sec:act. constraint}
Our planner must respect the actuator constraints. If we refer to eq.~(\ref{eqn:thrust}), then it follows that there are lower and upper bounds $\tau_{min}$ and $\tau_{max}$ respectively that the vehicle's thrust $\tau$ should respect. We can then incorporate the actuator constraint formulated as
\begin{equation}\label{eqn:global constraint}
\begin{split}
\tau_{min}^2 \leq \left\lVert m\Ddot{\bm{x}}+mg \mathbf{e}_3 \right\rVert^2_2 \leq\tau_{max}^2.
\end{split}
 \end{equation}
Since this constraint is nonlinear, it cannot be formulated in the QP optimization. We will describe how our system satisfies this condition in the next section. 
  
 \subsubsection{Optimization Procedure} 
\begin{algorithm}[!t]
\caption{{\sc Global Bound Checking (GBC)} }
\label{agl:bound}
Returns true if $H\left(t\right) < b$ $\forall$ $t\in\left[t_0,t_f\right]$. $H(t)$ is any polynomial. $b \in \mathbb{R}$ is an upper bound
\begin{algorithmic}[1]
    \STATE Let $F(t) = H(t)- b$;
    \IF{$F\left(t_0\right) > 0 \hspace{10pt} \OR\hspace{10pt} F\left(t_f\right) > 0 $} 
        \RETURN FALSE
    \ENDIF
    \IF{STURM$\left(F(t), t_0,t_f\right) > 0 $} 
        \RETURN FALSE
    \ENDIF
    \RETURN TRUE
 \end{algorithmic}
\end{algorithm}

For our planner, we use the QP to solve the constraints described in eqs.~(\ref{eqn:acceleration constraint}),~(\ref{eqn:Velocity constraint}), and~(\ref{eqn:acceleration discrete ineq constraint}).  Next, we use Algorithm~\ref{agl:bound} to verify if the constraint described in eq.~(\ref{eqn:global constraint}) is met. We do not apply Algorithm~\ref{agl:bound}  to the inequality constraints in eqs.~(\ref{eqn:Velocity constraint}) and (\ref{eqn:acceleration discrete ineq constraint}) because they are linear constraints and can be efficiently resolved within the QP optimization. Should the global bounds be violated, we increase the trajectories' time iteratively and recursively solve the QP until eq.~(\ref{eqn:global constraint}) is met. We visualize this process in Fig.~\ref{fig:GBC} whereby increasing the allotted time for a minimum snap trajectory, the constraint from eq.~(\ref{eqn:global constraint}) is met through iteratively expanding the time and checking the bounds.

Inspired by~\cite{wang2020alternating},  Algorithm \ref{agl:bound}  checks if given any polynomial, $H(t)$, then is $H(t) < b \hspace{10pt} \forall \hspace{10pt} t \in \left[t_o,t_f\right]$ true. For our algorithm, $H(t)$ is set to the squared norm of the thrust described in eq.~(\ref{eqn:global constraint}). Algorithm \ref{agl:bound} works by leveraging the function STURM~\cite{sturm2009} which returns the number of roots of any arbitrary polynomial, $H(t)$, in a bound $t \in\left[t_0,t_f\right]$. A proof of algorithm~\ref{agl:bound} is included in the appendix along with a more detailed description of STURM's implementation. Using algorithm~\ref{agl:bound}, we can ensure that our generated trajectory does not violate eq.~(\ref{eqn:global constraint}) without checking each point of our trajectory. This time reduction is quantified in Table.~\ref{tab:computation}.




\section{Experimental Results}~\label{sec:experiment_results}
The experiments are conducted in the new indoor testbed with a flying space of $10\times5\times4~\si{m^3}$ at the ARPL lab at New York University. The ground truth data is collected using a Vicon\footnote{\url{www.vicon.com}} motion capture system at $100~\si{Hz}$.  Navigation is performed solely based on the on board VIO system running at 500 Hz. Our landing pad is mounted on an adjustable desk that allows us to control height. Also, the adhesive is attached to an adjustable stand which allows us to control the surface angle along a specific axis. We selected a tolerance, $q=0.1$, sampling time, $dt=0.01~\si{s}$, time before impact, $t_k = 0.15~\si{s}$ and $\alpha = 3.3~\si{m/s^2}$ as the hyperparameters of eqs.~(\ref{eqn:acceleration constraint}) and (\ref{eqn:acceleration discrete ineq constraint}). We choose to minimize the $j=4$ ,snap norm,  for eq.~(\ref{eqn:cost_integral_QP}). Finally, we set the impact velocity bound of eq.~(\ref{eqn:Velocity constraint}) as the minimum impact velocity, $v_{min} =0.4 \si{m/s}$, and maximum impact velocity, $v_{max}=0.6 \si{m/s}$. 

\begin{figure}[!t]
    \centering
    \includegraphics[width=\linewidth,height=130pt]{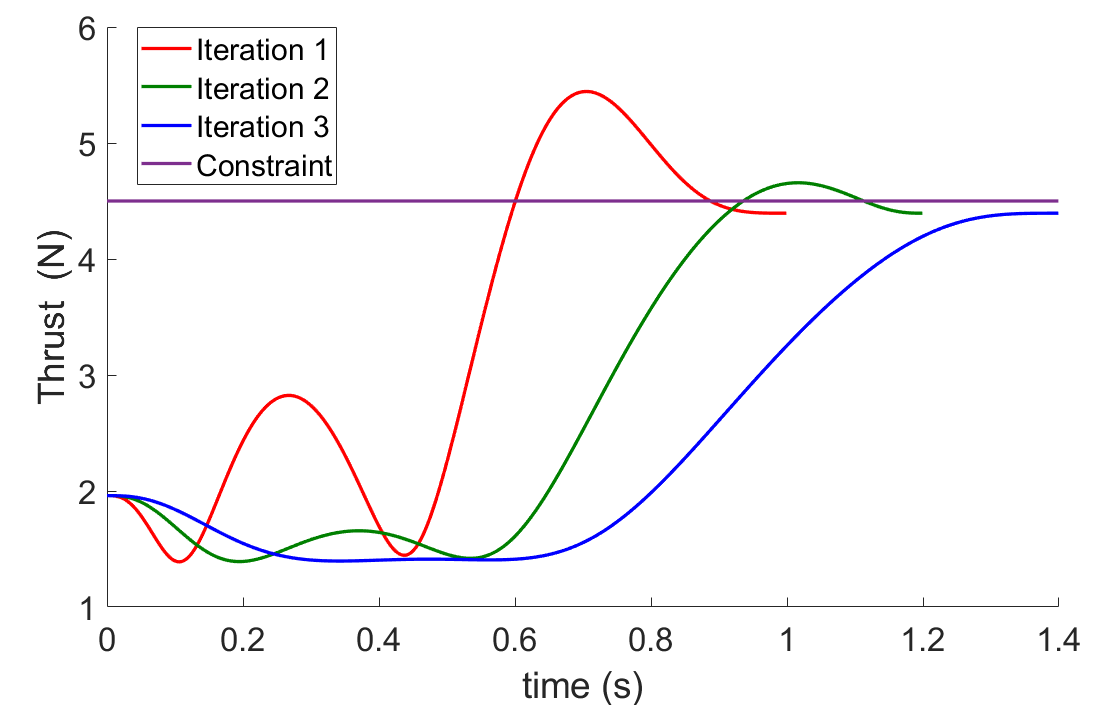}
    \caption{Iterations through the GBC reducing thrust for perching trajectory.}
    \label{fig:GBC}
    \vspace{-20pt}
\end{figure}

First, we verify that our optimization procedure respects the maximum thrust bound in addition to the linear inequality constraints. This trajectory  times spanned an initial $t_0=0~\si{s}$ and end $t_f=1~\si{s}$. We then apply a bound on the maximum thrust, $\tau_{max} =4.5 ~\si{N}$. Visualized in Fig.~\ref{fig:GBC}, the thrust shrinks for each iteration as time increases. This process repeats till our condition is respected, where $t_f = 1.4~\si{s}$.

Next, we evaluate the computational time and scalability of our trajectory generator by analyzing the time taken to resolve a $3$ and $10$ spline problem. Computational results are reported in Table~\ref{tab:computation}. To ensure consistency of our results, we obtained the results by re-running the optimizer $5$ times and averaging the outcome. We leverage the C++ OOQP library~\cite{OOQP} to solve our QP formulation. For the following experiment, we compared whether using one iteration of the nonlinear optimizer package NLOPT~\cite{NLOPT} to guarantee thrust, or checking and reformulating multiple iterations of the QP problem was faster. The following experiments all use a polynomial order $N=14$.
We provide results for this specific scenario in Table~\ref{tab:computation}. The nonlinear optimization takes on the order of a $100$ times longer to solve than performing a QP algorithm and repeating a check afterwards. This indicates that our current approach of doing multiple QP optimizations is better than trying to solve a single more complex nonlinear optimization.  
\begin{table}[!t]
\centering
\resizebox{1\columnwidth}{!}{ 
 \begin{tabular}{|c|c|c|c} 
 \hline
 Optimizer & Num. Waypoints & Computation Time (ms) \\ [0.5ex] 
 \hline
 \multirow{2}{*}{QP+GBC} & 3 & 4 \\
                         & 10 & 18\\
                     \hline
 \multirow{2}{*}{NLOPT} & 3 & 527\\ 
                        &10  & 1253 \\
\hline
\end{tabular}
}
\caption{Computation time of our approach as a function of the number of waypoints.}
\label{tab:computation}
    \vspace{-10pt}
\end{table}

\begin{figure*}[!t]
    \centering
     \includegraphics[width=\linewidth,,height=110pt]{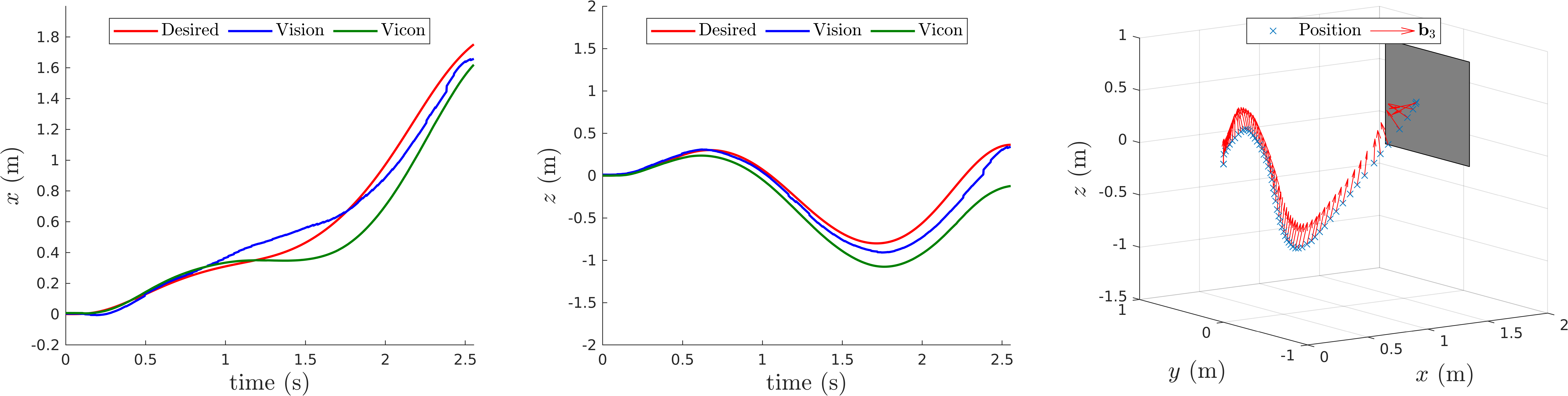}
    \caption{Trajectory tracking and localization results for $90^{\circ}$ surface inclination from a distance of $1.7~\si{m}$. The blue crosses represent the quadrotor position, whereas the arrows represent the thrust vector.}
    \label{fig:Tracking}
\end{figure*}

\begin{figure}[!ht]
    \centering
    \includegraphics[width=\linewidth,height=110pt]{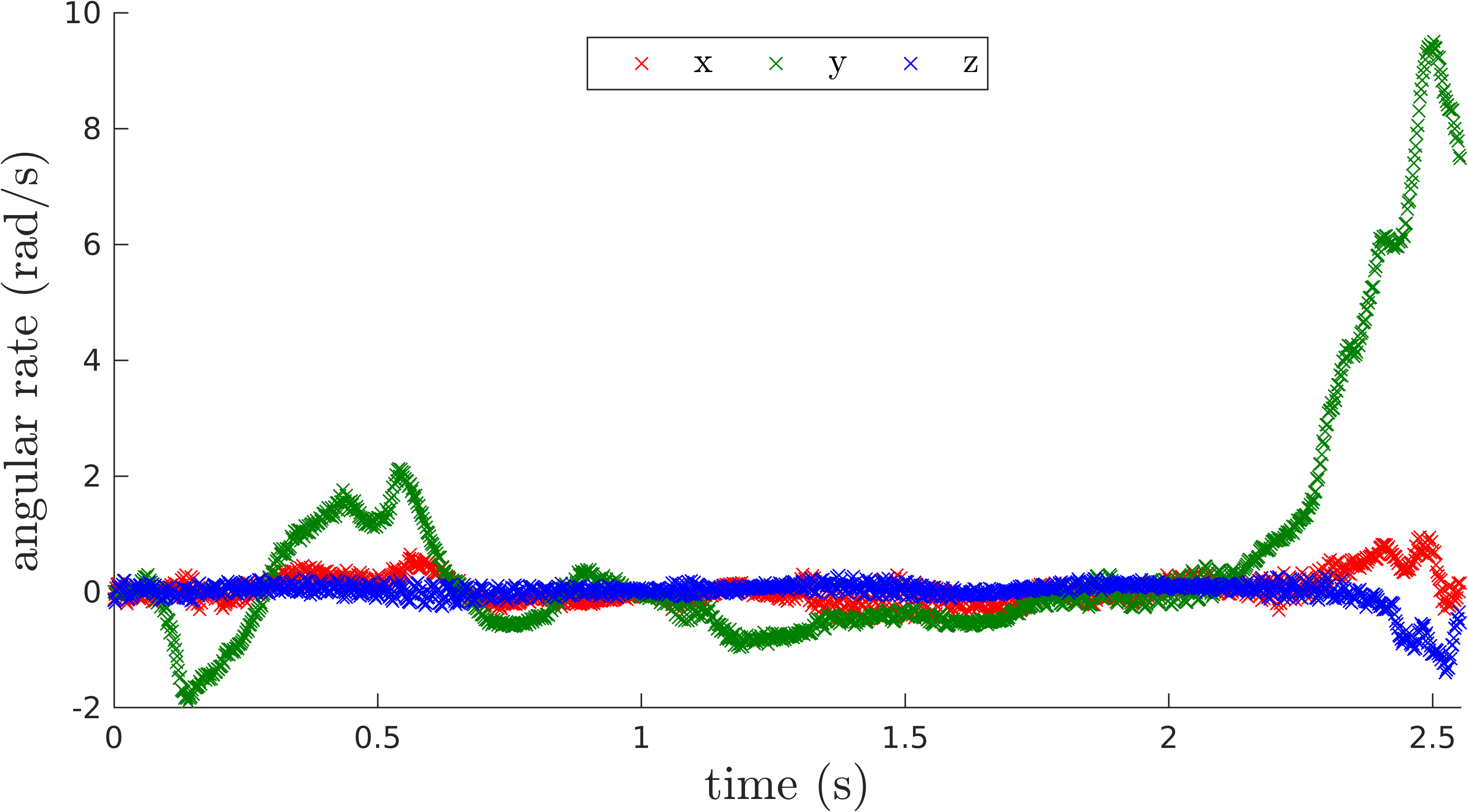}
    \caption{Angular rate during a $90^\circ$ perching maneuver.}
    \label{fig:angular_rate}
\vspace{-10pt}
\end{figure}

We then proceed to evaluate our approach for perching on an inclined surface as shown in Fig.~\ref{fig:90_perchingimage} considering two challenging inclination angles of $60^\circ$ and $90^\circ$. In Fig.~\ref{fig:Tracking}, we present the trajectory planning, control tracking, and localization results for the most challenging perching maneuver at $90^\circ$. We do not report the results on y-axis in Fig. \ref{fig:Tracking} because the motion along that axis does not have significant variations during the perching task. We also plot the thrust vector $\mathbf{b}_3$ in Fig.~\ref{fig:Tracking} (right) to further show that the constraint during perching is correctly enforced during the execution of the maneuver. As demonstrated, our additional constraints imposed through the eqs.~(\ref{eqn:acceleration constraint}) and (\ref{eqn:acceleration discrete ineq constraint}) enforce the rotation to complete slightly before reaching the target. 
 To further show that the proposed maneuver is aggressive and challenging, we present the angular rates achieved during $90^{\circ}$ perching on the three Cartesian axes in Fig~\ref{fig:angular_rate}. Our vehicle achieves angular rates close to $600 ~\si{deg/s}$, which to the best of our knowledge has never been achieved in the past for a small scale vehicle using on board computation and visual perception for both state estimation and target localization.

\begin{table}[!t]
\centering
\begin{tabularx}{0.48\textwidth}{c c c c c c}
    \hline\hline
 \rule{0pt}{2ex} &Component&\multicolumn{2}{c}{1.7 m}&\multicolumn{2}{c}{3 m}\\
  \cline{3-6}\
  \rule{0pt}{2ex}  & & Tracking    & Estimation  & Tracking & Estimation\\\hline
  $60^\circ$&x$~(\si{m})$&0.0500   & 0.0145      & 0.0378   & 0.0401 \\
            &y$~(\si{m})$&0.0325   & 0.0851      & 0.0732   & 0.0714 \\
            &z$~(\si{m})$&0.1509   &  0.0714     & 0.1321   & 0.0452 \\\hline
  $90^\circ$&x$~(\si{m})$&0.0503   & 0.0314      & 0.0948   & 0.0660 \\
            &y$~(\si{m})$&0.0539   & 0.0533      & 0.0360   & 0.0045 \\
            &z$~(\si{m})$&0.1081   & 0.0405      & 0.0911   & 0.0485 \\\hline
    \hline\hline
\end{tabularx}
\caption {Tracking and Estimation RMSE for different surface inclinations and distances.\label{tab:tracking_error}}
\vspace{-10pt}
\end{table}

Finally to validate the consistency, performance, and robustness of our algorithm, we use $2$ different surface inclinations at two target distances of $1.7~\si{m}$ and $3~\si{m}$ respectively. $3~\si{m}$ is the maximum distance to reliably detect the Apriltags on our quadrotor. Table~\ref{tab:tracking_error} shows the relevant trajectory tracking and state estimation RMSE metrics of these experiments. Furthermore, we repeat the procedure $5$ times for each distance and surface inclinations and record the successful perching rate as seen in Table~\ref{tab:success_rate}. A successful perching is defined as the quadrotor adheres to the target and remains attached. The lower success rate at $60^\circ$ is due to the Apriltag accuracy falling for a worse viewing angle and further target. 
The tracking error is mostly located toward the end of the trajectory as depicted in Fig.~\ref{fig:Tracking}. Once the vehicle has angular velocity prior to perching, it is very difficult to control its position.
However, we still see our controller succeeding in the perch in these conditions. Overall, we see that our approach can generate reliable perching maneuver based on on board perception with cm level accuracy both in term of localization and control tracking of the maneuver. Furthermore, these success rates and tracking errors are agnostic with respect to the tested starting positions as well as surface inclination.   

\begin{table}[!t] 
\centering
\resizebox{0.7\columnwidth}{!}{  
\begin{tabular}{|c|c|c|}
\hline
 Target  & Surface & Success \\
 Distance (m)  & Angle &  rate\\
 \hline
 \multirow{2}{*}{$1.7$} 
       & $60^\circ$   & 4/5 \\ 
       & $90^\circ$ & 4/5 \\
 \hline
   \multirow{2}{*}{$3$} 
       & $60^\circ$   &  3/5\\
       & $90^\circ$ & 4/5\\
 \hline
\end{tabular}
}
\caption{Success rate statistics as a function of the inclined surface and the target distance.}\label{tab:success_rate}
    \vspace{-15pt}

\end{table}

\section{Conclusion}~\label{sec:conclusion}
In this paper, we presented the planning, control, and perception methodologies to achieve autonomous visual perching with small quadrotors relying exclusively on on board computation and sensing. Our results show that we can generate aggressive and challenging perching maneuvers up to 90$^\circ$ inclined surfaces, angular rates up to  $600~\si{deg/s}$, and accelerations up to $10~\si{m/s^2}$. These results show the agility and robustness of our real-time autonomous perching.

Future works will focus on two research directions. First, we aim to leverage consecutive target detection across the entire planned maneuver to have a receding-horizon planning strategy. Continuous target information introduces the possibility to refine the trajectory in real-time to compensate for possible drifts or unmodelled effects during flight and consequently increases the resilience and precision for target interception. In this context, it is interesting to study the trade-off among the dynamic feasibility, aggressive behavior of the perching maneuver, and maximizing the visibility of the target. Second, we will work on enforcing the planning objectives and constraints at the control level by formulating a nonlinear model predictive control problem where the sensors and actuator constraints can be embedded as additional terms in the optimization cost function or as constraints.




\section*{Appendix}
\subsection{Sturm's Theorem}
Sturm's theorem~\cite{sturm2009} states that the number of roots for a polynomial, $H\left(t\right)$ in an interval $\left[t_0,t_f\right]$ is equal to the difference in sign changes of the Sturm's sequence, eq.~(\ref{eqn:sturm_seq}), between $S\left(t_0\right)$ and $S\left(t_f\right)$  
\begin{equation}\label{eqn:sturm_seq}
    S\left(t\right) =  \begin{cases}
       S_0\left(t\right) =H\left(t\right) \\
       S_1\left(t\right) =\dot{H}\left(t\right)\\
        S_{i+1}\left(t\right)=  -Rm\left(S_{i-1},S_i\right) \\
        \vdots\\
        S_{N}\left(t\right) = -Rm\left(S_{N-2},S_{N-1}\right) \in \mathbb{R}
     \end{cases},
\end{equation}
where $Rm(S_{i-1}, S_i)$ gives the algebraic remainder of $\displaystyle\frac{S_{i-1}}{S_{i}}$.


\subsection{Proof of Algorithm 1}
First, we leverage the intermediate value theorem. The intermediate value theorem state that given a continuous function $H\left(t\right)$ whose domain contains the values $\left[t_0,t_f\right]$ then $\forall \hspace{10pt} i \in \left[H(t_0), H(t_f)\right]$ there must exist a corresponding $t_i\in\left[t_0,t_f\right]$ such that $i = H(t_i)$. Since our trajectory is continuous, this theorem holds in our case. Now let's prove that our algorithm works by contradiction. Assume, there exists a $t_i\in\left[t_0,t_1\right]$ such that $H\left(t_i\right) > b$ where $b$ is the global bound, and all conditions of Algorithm 1, $H\left(t_0\right) < b$, $H\left(t_f\right) < b$, and $H\left(t_i\right)-b \neq 0~\forall~t_i\in \left[t_0,t_f\right]$ are true.

If this is the case, then we can apply the intermediate value theorem and construct a domain $\left[t_0, t_i\right]$ and a range $\left[H\left(t_0\right), H\left(t_i\right)\right]$.  We know that $H\left(t_0\right) < b$ and $H\left(t_i\right) > b$, then $b \in \left[H\left(t_0\right), H\left(t_i\right)\right]$. Therefore, based on the intermediate value theorem, there must exist a $t_j \in \left[t_0,t_i\right]$ such that $H\left(t_j\right)=b$. However, we see $H\left(t_j\right)=b$ is a contradiction with respect to the condition $H\left(t_j\right)-b \neq 0~\forall~t_j\in \left[t_0,t_f\right]$. As $\left[t_0,t_i\right] \subset \left[t_0,t_f\right]$ by construction, this condition also holds true for all $t_j \in \left[t_0,t_i\right]$. Since, this is a contradiction there can exist no such number $t_i \in \left[t_0,t_1\right]$ such that $H\left(t_i\right) > b$ if our algorithm returns true.

\addtolength{\textheight}{-12cm}   

\bibliography{ref}
\bibliographystyle{IEEEtran}

\end{document}